%% file: root.tex
\documentclass[letterpaper, 10 pt, conference]{ieeeconf}  % Comment this line out if you need a4paper

\IEEEoverridecommandlockouts                              % This command is only needed if 
\usepackage{graphics}
\usepackage{amsmath}
\usepackage{amssymb}
\usepackage{balance}
\usepackage{mathtools}
\usepackage{hyperref}
\usepackage{xcolor}
\usepackage[font=footnotesize,hypcap=false]{caption}
\usepackage{siunitx}

\usepackage{subcaption}

\title{\LARGE \bf
Learning Bipedal Walking for Humanoids with Current Feedback
}

\author{Rohan P. Singh$^{1,2}$, Zhaoming Xie$^{3}$, Pierre Gergondet$^{1}$, Fumio Kanehiro$^{1,2}$% <-this % stops a space
\\\tt\small{Email: rohan-singh@aist.go.jp}
\thanks{
$^{1}$
CNRS-AIST JRL (Joint Robotics Laboratory) IRL,
National Institute of Advanced Industrial Science and Technology (AIST),
Japan.}
\thanks{
$^{2}$
University of Tsukuba,
Ibaraki,
Japan.}
\thanks{
$^{3}$
Department of Computer Science,
Stanford University,
USA.}
}%

\begin{document}

\maketitle
\thispagestyle{empty}
\pagestyle{empty}

%%%%%%%%%%%%%%%%%%%%%%%%%%%%%%%%%%%%%%%%%%%%%%%%%%%%%%%%%%%%%%%%%%%%%%%%%%%%%%%%
\input{sections/abstract}
\input{sections/intro}
\input{sections/related}
\input{sections/background}
\input{sections/approach}
\input{sections/experiments}
\input{sections/conclusion}
\input{sections/acknowledgements}

\balance
\bibliographystyle{IEEEtran}
\bibliography{IEEEabrv,bibliography.bib}
\end{document}

%% file: sections/abstract.tex
\begin{abstract}
Recent advances in deep reinforcement learning (RL) based techniques combined with training in
simulation have offered a new approach to developing robust controllers for legged robots.
However, the application of such approaches to real hardware has largely been limited to quadrupedal
robots with direct-drive actuators and light-weight bipedal robots with low gear-ratio transmission
systems. Application to real, life-sized humanoid robots has been less common arguably due to a large
\textit{sim2real} gap.

In this paper, we present an approach for effectively overcoming the \textit{sim2real} gap
issue for humanoid robots arising from inaccurate torque-tracking at the actuator level.
Our key idea is to utilize the current feedback from the actuators on the real robot, after
training the policy in a simulation environment artificially degraded with poor torque-tracking.
Our approach successfully trains a unified, end-to-end policy in simulation that can be deployed on a real
HRP-5P humanoid robot to achieve bipedal locomotion.
Through ablations, we also show that a feedforward policy architecture combined with targeted
dynamics randomization is sufficient for zero-shot \textit{sim2real} success, thus eliminating
the need for computationally expensive, memory-based network architectures.
Finally, we validate the robustness of the proposed RL policy by comparing its performance against a
conventional model-based controller for walking on uneven terrain with the real robot.
%
% We release the code for training and evaluation is publicly online \footnote{\url{https://github.com/rohanpsingh/LearningHumanoidWalking}}.
\end{abstract}

%% file: sections/intro.tex
%%%%%%%%%%%%%%%%%%%%%%%%%%%%%%%%%%%%%%%%%%%%%%%%%%%%%%%%%%%%%%%%%%%%%%%%%%%%%%%%
\section{Introduction}
\label{section:intro}

%%%%%%%%%%%%%
%%%%%%%%%%%%%
\begin{figure}[t]
  \includegraphics[width=\linewidth]{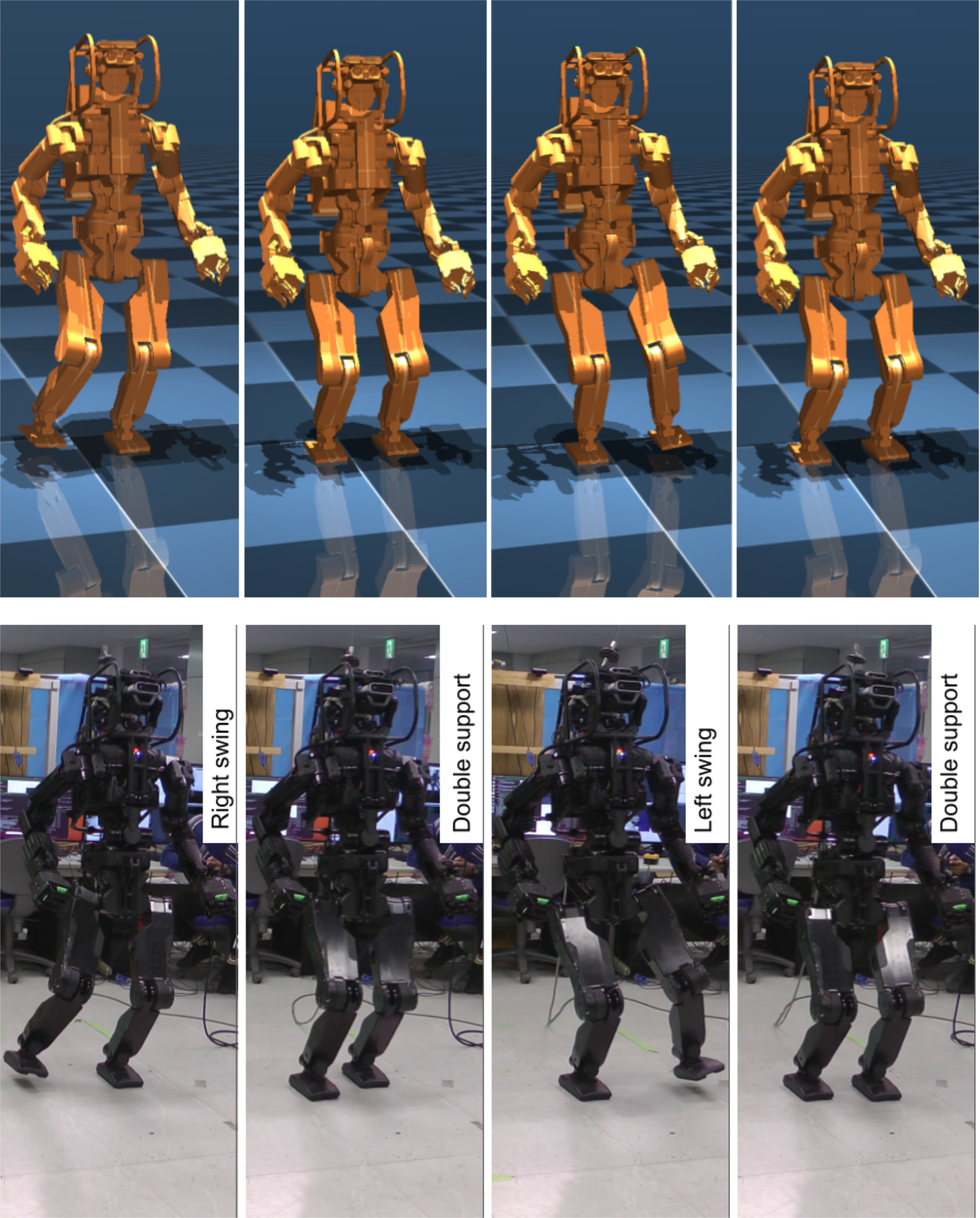}
  \caption{\textbf{HRP-5P humanoid robot} trained to perform bipedal locomotion via model-free
    reinforcement learning in MuJoCo (top); RL policy transferred to the real robot (bottom). We
    make use of the feedback from measured actuator current to account for the poor torque-tracking
    on the real system.
    }
  \label{figure:intro}
\end{figure}
%%%%%%%%%%%%%
%%%%%%%%%%%%%

As conventional model-based approaches for humanoid locomotion continue to improve,
such as those based on preview control \cite{murooka2022centroidal} or model predictive control (MPC)
\cite{romualdi2022online}, their robustness against unexpected disturbances and inaccurate modeling is
still an elusive research goal. On the other hand, rapid advancements in RL-based control
methods for legged locomotion have shown outstanding performance in unstructured
and uncontrolled environments for quadrupedal robots
\cite{2022-rss-quadruped, 2019-scirobotics-anymal_learning,2021-rss-RMA,2020-science-blindQuadruped}
and even bipedal robots \cite{2019-CORL-cassie, siekmann2021sim, siekmann2021blind}.
It would be appealing to apply similar methods to develop walking controllers for
larger and heavier humanoid robots, too.

Training a capable policy using deep RL is data intensive and can be damaging to the hardware.
Physics simulation environments offer a safe way to collect a large amount of data, so policies are typically
trained in simulation and then transferred to the real system.  
However, the simulated environment can fail to capture the richness of real-world dynamics.
This gives rise to the ``reality gap'', more commonly known as the \textit{sim2real} gap.
The \textit{sim2real} gap can cause the performance of a policy trained in simulation to drop drastically
when deployed on the real hardware. 
In the case of life-sized humanoid robots such as the HRP-series humanoids, this gap can have a more
critical effect on the robot's stability during walking, compared to the quadrupedal robots or lightweight bipedal
robots that are used in most of the recent works.
Memory-based policy architectures that can use temporal information to essentially perform online system
identification have previously been proposed to tackle this issue \cite{2020-rss-RNNCassie, 2021-rss-RMA}.
But such networks are generally more computationally expensive than feedforward (FF) networks, which can be
prohibitive for prototyping and deployment of RL policies especially for learning tasks with high sample
complexity.

Previously, it has been reported that the simulated actuator model has a major impact on the \textit{sim2real}
transfer compared to other factors such as link masses and their center-of-mass (CoM) positions
\cite{2019-scirobotics-anymal_learning, 2019-iros-sim2realBiped, masuda2022sim}. We argue that within
the actuator model, the modelling of the torque-tracking behavior forms a key source of discrepancy between the
real and simulated robots. In the simulation environment, the joint torque desired by the RL policy is injected
exactly into the dynamics computation as the control input. However, in the case of real robots such as the
current-controlled HRP-5P, the actual torque exerted by the motor on the link may be significantly different
from the desired torque due to imperfect current-control (and a nearly linear relationship between torque and current for brushed DC motors).
Since the robot makes environmental interaction only by applying torques on the links,
the mismatch between the torque desired by the policy and the actual torque applied on the link may have
consequential effects on the control.

%% Furthermore, these robots have heavy limbs to support a heavy upper body, which requires a high gear-ratio transmission system
%% with a large output torque range. Consequently, the actuators used have relatively large armature
%% (also known as rotor inertia) and low backdrivable joints. High gear-ratio also induces
%% other hard-to-model but non-trivial effects such as joint friction and back-EMF. This makes the \textit{sim2real} gap for a robot such as HRP-5P (\autoref{figure:intro}) much larger than the lighter and more backdrivable bipedal systems like Cassie and Digit.
%% Hence, the suitability or relevance of recent \textit{sim2real} successes in the literature for this robot is a matter of investigation.

In this paper, we develop a system to train bipedal walking policies in simulation and deploy them on the HRP-5P
humanoid robot (\autoref{figure:intro}). HRP-5P is a high-power, electrically-actuated, 53 degrees of freedom (DoF)
humanoid robot weighing 105kg, with a height of $182cm$ \cite{kaneko2019humanoid}. Our key insight is that, on such
robots, the \textit{sim2real} gap is significantly a result of imperfect current tracking on the real robot that leads
to a mismatch in the desired and applied torque. To this end, we propose to simulate a degraded torque-tracking effect during training
and to incorporate current feedback from the motors into the observation space. The resulting policy learns to actively use
the current feedback signal and compensate for the inaccurate torque-tracking
within the motor drivers. Finally, the policy could be successfully deployed on a real HRP-5P robot.

Specifically, our contributions in this work are as follows:
\begin{itemize}
  \item{}
    By using the proposed \textit{sim2real} approach, we show one of the first demonstration of an end-to-end policy for
    a real, current-controlled, humanoid robot to achieve dynamic stability. Our policy can achieve
    forward walking, stepping and turning in-place, and quite standing, by receiving user commands
    via a joystick.
  \item{}
    We perform ablation study on feedforward Multi-Layered Perceptrons (MLPs) and Long Short Term Memory (LSTMs) networks to
    show that it is possible to bridge the ``reality gap'' without relying on memory-based policy architectures
    or resorting to unreasonably wide randomization of dynamics parameters. This is necessary for the development of RL
    policies within reasonable amount of computation resources.
  \item{}
    We validate the robustness of the proposed policy on the real robot and compare the qualitative performance of
    the RL policy to an open-source model-based walking controller for blind locomotion over small uneven and compliant
    obstacles.
\end{itemize}
%%%%%%%%%%%%%%%%%%%%%%%%%%%%%%%%%%%%%%%%%%%%%%%%%%%%%%%%%%%%%%%%%%%%%%%%%%%%%%%%

%% file: sections/related.tex
\section{Related Work}
\label{section:related}

\subsection{Reinforcement Learning for Legged Robots}
Reinforcement Learning has become a powerful approach for synthesizing
controllers for legged robots. Control policies are typically trained
in simulation and then transferred to the hardware, i.e., \textit{sim2real}.
A large number of such works focus on quadrupedal robots, e.g.,
ANYmal \cite{2019-scirobotics-anymal_learning}, Laikago \cite{2020-rss-Peng},
A1 \cite{2021-rss-RMA}, Jueying \cite{2020-scirobotics-quadruped} and
Mini-Cheetah \cite{2022-rss-quadruped}. There are also successes in applying
the same approach for bipedal robots, e.g., on the Cassie
\cite{2019-CORL-cassie, 2020-rss-RNNCassie, 2022-IROS-BerkeleyCassieRL},
Digit \cite{2022-IEEE-DigitRL, 2021-IROS-DigitRL, radosavovic2023learning} and NimbRo-OP2X \cite{2021-ICRA-DeepWalk}.
DeepWalk \cite{2021-ICRA-DeepWalk} demonstrates a single learned policy for a real
humanoid robot that can achieve omnidirectional walking on a flat floor through the
use of Beta policies \cite{chou2017improving}, albeit noting a need for more
sophisticated methods for improved transfer.

For the HRP-5P and JVRC-1 humanoids, end-to-end deep RL policies have been
previously been demonstrated for walking on planned footsteps, but only in the
simulation environment \cite{singh2022learning}. The focus of this work, on the
other hand, is to solve the \textit{sim2real} issues for achieving real robot demonstrations.

Recently, an impressive demonstration of torque-based deep RL locomotion policy
for the real TOCABI humanoid robot has shown the advantage of using torque
action-space for improved \textit{sim2real} transfer \cite{kim2023torque}.
TOCABI is a human-sized humanoid \cite{schwartz2022design} that can achieve
torque-controlled compliance without the use of joint-level torque sensor.
Application of such an approach for the low backdrivable joints of HRP-5P
humanoid is a matter of future research.

Again, implementing learned policies on the hardware of a bulky humanoid robot,
such as Valkyrie and the HRP-series robots, presents significantly greater
challenges than lighter robots, especially due to heightened safety risks.

\subsection{Sim2Real Approaches}
Important ingredients for successful \textit{sim2real} include
(a) careful system identification, e.g, a learned actuator model is incorporated in the simulator to account for hard-to-model actuator dynamics \cite{2019-scirobotics-anymal_learning, 2019-iros-sim2realBiped},
(b) domain adaptation, where the policy learns to adapt based on the history of observations and actions \cite{2021-rss-RMA, 2020-science-blindQuadruped, 2020-rss-RNNCassie},
(c) dynamics randomization, where simulation parameters such as mass, friction, inertia, CoM positions
are randomized to improve robustness of the policy \cite{2018-rss-sim2realQuadruped}.
While successful on several legged robot platforms, we are yet to see such
approaches being successfully applied to life-size (e.g., more than 170cm)
humanoid robots with heavy limbs. This is not due to lack of attempt,
prior work has explored how to apply reinforcement learning to the NASA
Valkyrie robot, e.g.,~\cite{2018-humanoids-ValkerieRL}, but so far no
\textit{sim2real} is demonstrated.

For the case of HRP-5P, end-to-end supervised learning of actuator dynamics
\cite{2019-scirobotics-anymal_learning} is not readily applicable due to absence
of joint torque sensors which could provide the ground-truth signal for such an
actuator model. Further, search-based system identification methods using real
data \cite{2019-iros-sim2realBiped, masuda2022sim} may be difficult for a large
robot as they require multiple trial runs (5-10) on the real robot.
Domain adaptation and online system identification methods through the use of
memory-based agents have been demonstrated successfully for biped robots
\cite{2020-rss-RNNCassie, 2021-rss-RMA}. However, training memory-based networks
(such as LSTMs) can be significantly more computationally expensive and may
prohibit real-time inference on the real robot. Our experiments show that with the
combination of targetted dynamics randomization, we can bridge the ``reality gap''
using light-weight FF networks (experimental results in \autoref{subsec:policy-arch}).

\subsection{Conventional Control for Humanoid Locomotion}
Conventional model-based approaches for humanoid bipedal locomotion consist of
local feedback controllers to track Zero-Moment-Point (ZMP) or Center-of-Mass (CoM)
trajectories precomputed in an offline process.
Stabilization control through the use of divergent component of motion (DCM) has been
extensively studied in prior works \cite{englsberger2015three, caron2019stair} and
applied to real robots. It relies on the linear inverted pendulum model \cite{kajita2003biped, kajita2010biped}
of bipedal walking. Other recent works such as \cite{murooka2022centroidal}, on the other
hand, do not rely on biped-specific dynamics and instead perform online generation
of centroidal trajectory based on preview control for impressive multi-contact motion.
The method uses a preview control scheme to generate a centroidal trajectory and
a stabilization scheme to correct errors in tracking the trajectory. Since
the trajectory is generated online the robot can react robustly to environmental
disturbances.

Generally, the output of such controllers is in the form of desired ZMP, CoM, and/or
contact wrenches, which are then fed to Quadratic programming (QP) solver to compute
desired joint positions. The desired joint positions are then tracked by a local
proportional-derivative (PD) controller for stiff position control. This is in sharp contrast to RL approaches
mentioned above that use low PD gains to achieve compliant tracking, and consequently,
offer greater robustness to uneven terrain.

To this end, we perform real robot experiments to compare the robustness of our proposed
RL policy to a commonly used, open-source model-based controller for locomotion over
uneven terrain (details and results provided in \autoref{subsec:robustness-tests}).

%% file: sections/background.tex
%%%%%%%%%%%%%%%%%%%%%%%%%%%%%%%%%%%%%%%%%%%%%%%%%%%%%%%%%%%%%%%%%%%%%%%%%%%%%%%%
\section{Background}

\begin{figure*}[t]
  \includegraphics[width=\linewidth]{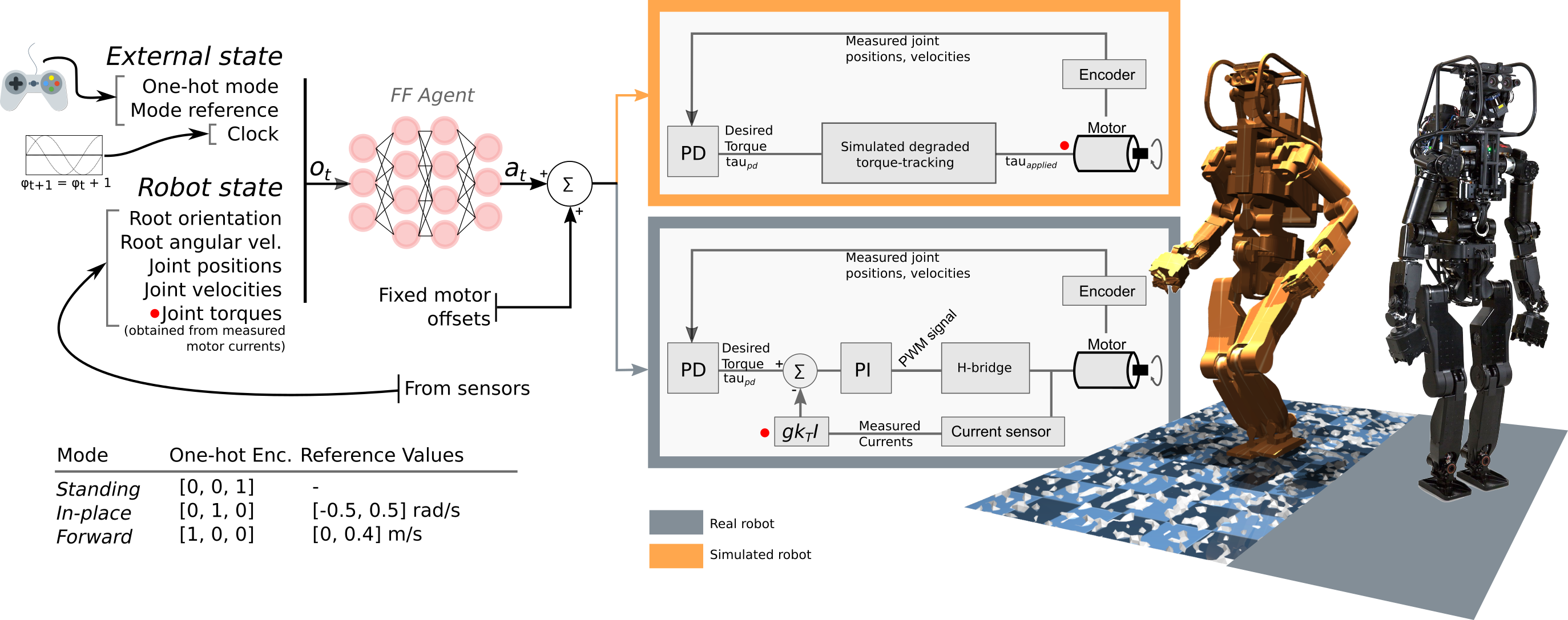}
  \caption{
    \textbf{Overview of the proposed RL policy with block-diagram of motor control system in HRP-5P.}
    Output of the RL policy in terms of ``desired joint positions'' is summed with fixed motor-offets
    (corresponding to the robot's nominal pose) and is fed to the joint PD controller. On the real robot,
    the torque computed by the PD controller is tracked by a PI current controller,
    albeit, with significant tracking errors. These tracking errors form a crucial
    component of the \textit{sim2real} gap. To overcome this issue, we propose to (a) simulate a degraded
    torque-tracking environment during training and (b) observe the applied motor
    torque, or equivalently, measured motor currents on the real robot (denoted by \textcolor{red}{\ensuremath\bullet}).
  }
  \label{figure:framework}
\end{figure*}

The motor control system on the HRP-5P robot (and generally, on most robot
transmission systems) is shown in \autoref{figure:framework}. The block
structure consists of a PD controller which computes
the desired torque command given the reference position from a higher-level
controller and measured position from the joint encoder. The desired torque
command --- or equivalently, the current command (assuming a proportional relationship
between torque and current for a brushed-DC motor) is then sent to a proportional-integral
(PI) controller. The PI controller tries to track the current command given the
measured current from the current sensor in the motor. The output of the PI controller
is fed to the motor power amplifier, which in turn drives the motor.

The key observation here is that the PI controller is unable to precisely track the
torque commands, as desired by the higher-level controller or policy, often leading
to significant torque-tracking errors. We suspect that such tracking errors could be
caused largely due to the effect of the back electro-motive force (EMF). When the
motor rotates, the back-EMF creates a counter-voltage that opposes the applied voltage,
reducing the current flowing through the armature, leading to tracking errors. This
makes the real system vastly different from simulation environments where the desired
torque command is applied \textit{exactly} to the actuator without errors. Other factors
such as the battery voltage, resistance of the transmission cables, changes in load, or
poorly tuned gains of the PI controller may also contribute to poor current tracking.

%%%%%%%%%%%%%
%%%%%%%%%%%%%
%%%%%%%%%%%%%
%%%%%%%%%%%%%

%%%%%%%%%%%%%%%%%%%%%%%%%%%%%%%%%%%%%%%%%%%%%%%%%%%%%%%%%%%%%%%%%%%%%%%%%%%%%%%%

%% file: sections/approach.tex
%%%%%%%%%%%%%%%%%%%%%%%%%%%%%%%%%%%%%%%%%%%%%%%%%%%%%%%%%%%%%%%%%%%%%%%%%%%%%%%%
\section{Approach}
In this section, we detail each component involved in training the RL policy (see \autoref{figure:framework}).
The training is performed in the MuJoCo simulator \cite{todorov2012mujoco}.
In particular, we describe how we overcome the poor torque-tracking on the real robot
by simulating Back-EMF effect and using current feedback for the policy. Since the
current and torque on the robot's actuators are assumed to be proportional, we use
the terms interchangeably through the paper.

\subsection{Observations and actions}
The input to our policy consists
of the robot state, the external state, and a clock signal. The robot state includes the joint
positions and joint velocities of each actuated joint (6 in each leg), roll and pitch orientation
and angular velocity of the root (pelvis), similar to several related works \cite{singh2022learning, 2020-rss-RNNCassie}.
In this work, we propose to also include the motor-level torque signal for each actuator in the robot state $\text{tau}_{obs}$. In simulation, this
signal is equal to the actual torque $\text{tau}_{applied}$ applied to the actuators in the previous timestep. On the real robot,
$\text{tau}_{obs}$ needs to be derived from the raw current measurements, as explained in \ref{subsec:real-robot-fb}.
The external state vector comprises of a $3D$ one-hot encoding to denote the walking mode ---
$[0, 0, 1]$ for standing and $[0, 1, 0]$ for stepping in-place and $[1, 0, 0]$ for
walking forward. It also includes a $1D$ scalar which acts as a reference value depending on the mode:
If the active mode is \textit{Stepping}, the reference value denotes the turning speed; for \textit{Walking}
it denotes forward walking speed; and is ignored for the $\textit{Standing}$ mode.

The policy also observes a clock signal that depends on a cyclic phase variable $\phi$.
This variable is also used to define a periodic reward term in our reward function to generate walking behaviors.
We do a bijective projection of $\phi$ to a $2D$ unit cycle:

\begin{align}
\label{eq:clock}
\text{Clock} = \left\{ \sin \left( \frac{2\pi\phi}{L} \right), \cos \left( \frac{2\pi\phi}{L} \right)  \right\},
\end{align}

\noindent
where $L$ is the cycle period. $\phi$ increments from 0 to 1 at each control timestep and reset to 0
after every $L$ timesteps. Clock is then used as input to the policy.

The policy outputs desired positions of the actuated joints in the robot's legs. These predictions from
the network are added to fixed motor offsets corresponding to the robot's half-sitting posture. 
The desired positions are tracked using a low-gain PD controller, which computes the desired joint torque as follows:

\begin{align}
\label{eq:pdcontrol}
\text{tau}_{pd} = K_{p}(q_{des} - q) + K_{d}(0 - \dot{q}),
\end{align}

\noindent
where $K_{p}$ and $K_{d}$ denote the proprotional and derivate gain factors respectively. $q_{des}$ is
the policy prediction summed with the fixed motor offsets. $q$ and $\dot{q}$ denote the current joint
position and velocity.

\subsection{Reward function}
Our reward design ensures that a reference motion is not needed. Instead, we rely on several hand-crafted
reward terms that promote the desired robot behavior in 3 modes: stand in place, step in-place (including
turning) and walk forward given a reference speed. This requires
the robot to develop a periodic bipedal gait, follow the mode and reference velocity command and maintain a fixed
height. Further, we introduce terms to develop a more realistic motion that will allow \textit{sim2real}
transfer in a realistic and safe manner.

\textbf{Bipedal Walking.}
We introduce reward terms for promoting a symmetrical bipedal walking gait characterized by a periodic motion
of the legs, alternating between double-support (DS) phases, and the single-support (SS) phases. Depending on
the phase variable $\phi$ and the desired mode (\textit{standing} or \textit{walking}), the rewards for feet
ground reaction forces (GRF) and body speeds are computed.

For example, when $\phi$ lies in the first single-support region of the gait cycle (the left foot is in the
swing phase and the right foot is in the support phase), larger values of GRF on the left foot are rewarded negatively
while larger values of GRF on right foot lead to positive reward. Simultaneously, higher speeds for the left foot are
incentivized but penalized for the right foot.

The definition of the bipedal walking terms --- ground reaction forces at the feet $r_{\mathrm{grf}}$
and the feet body speeds $r_{\mathrm{spd}}$ --- are adopted exactly from \cite{singh2022learning}:

\begin{align}
& \text{$r_{\mathrm{grf}}$} = I_{left}^{grf}(\phi) \cdot F_{left} + I_{right}^{grf}(\phi) \cdot F_{right}
\label{eq:grf_reward} \\
& \text{$r_{\mathrm{spd}}$} = I_{left}^{spd}(\phi) \cdot S_{left} + I_{right}^{spd}(\phi) \cdot S_{right}
\label{eq:spd_reward}
\end{align}

\noindent
where $F_{left}$ and $F_{right}$ denote the GRF and $S_{left}$ and $S_{right}$ denote the body speeds on the
left and right foot respectively. We refer the reader to \cite{singh2022learning} for a detailed explaination of
the ``phase indicator" functions $I_{*}^{grf}$ and $I_{*}^{spd}$ for modulating the reward coefficients for
ground reaction forces and feet speeds.

For \textit{standing}, the DS phase is expanded to span the entire gait cycle, and the policy is rewarded
to maximize ground reaction forces on both feet while minimizing the feet speeds.

\textbf{Root Velocity, Orientation and Height.}
%% root fwd velocity term
The root linear velocity reward term is a simple cost on global speed $\prescript{x}{}{v_{root}}$ of the root link of the robot in the $x$-direction.
\begin{equation}
\label{eq:velocity_reward}
  r_{\mathrm{rv}} = \exp(-10 \cdot \| \prescript{x}{}{v_{root}} - \prescript{x}{}{\hat{v}_{root}} \| ^2)
\end{equation}

%% root yaw velocity term
The root yaw velocity term encourages the angular velocity of the root $\omega_{root}$ to be close to the desired
velocity $\hat{\omega}_{root}$.
\begin{equation}
\label{eq:yaw_vel_reward}
  r_{\mathrm{av}} = \exp(-10 \cdot \| \omega_{root} - \hat{\omega}_{root} \| ^2)
\end{equation}
\noindent

During training, the active mode is randomly selected between \textit{Standing}, \textit{Stepping}, and \textit{Walking}
at the start of an episode. Depending on the active mode, the scalar input for the reference value is sampled from a
uniform distribution, i.e., $\prescript{x}{}{\hat{v}_{root}}$ from a range of $[0, 0.4]\SI{}{\m\per\s}$
if mode is \textit{Walking} and $\hat{\omega}_{root}$ from a range of $[-0.5, 0.5]\SI{}{\radian\per\s}$
if mode is \textit{Standing}.

%% root height term
We also reward the policy to maintain the root height $h_{root}$ at a desired value $\hat{h}_{root} = \SI{0.79}{\m}$:
\begin{align}
\label{eq:height_reward}
  r_{\mathrm{height}} = \exp(-40\cdot(h_{root} - \hat{h}_{root})^2).
\end{align}

\textbf{Safe and realistic motion.}
In addition to the above terms, we also try to create a motion that remains close to the nominal posture of the robot
to avoid unnecessary sways. This is critical for safe deployment on the real robot, which has
a wide range of motion and significantly strong actuators.

%% upperbody term
To encourage the robot to maintain an upright posture, we use a reward term to minimize the distance
between the floor projection of the head position $\prescript{x,y}{}{\mathbf{p}_{head}}$ and the root position
$\prescript{x,y}{}{\mathbf{p}_{root}}$. This prevents the robot from developing a leaned-back behavior and excessively
swaying the upper body:
\begin{align}
\label{eq:upper_reward}
  r_{\mathrm{upper}} = \exp (-10 \cdot \|\prescript{x,y}{}{\mathbf{p}_{head}} - \prescript{x,y}{}{\mathbf{p}_{root}} \| ^2).
\end{align}

%% posture term
We use a term to penalize the distance of the current joint positions $\mathbf{q}$ from the nominal
``half-sitting posture'', $\mathbf{q_{nominal}}$:
\begin{align}
\label{eq:posture_reward}
  r_{\mathrm{posture}} = \exp (- \|\mathbf{q} - \mathbf{q_{nominal}} \| ^2).
\end{align}

%% joint velocity term
We also place a penalty on joint velocities $\mathbf{\dot{q}}$ above $50\%$ of the maximum
joint velocity $\mathbf{\dot{q}_{lim}}$.
\begin{align}
\label{eq:joint_vel_reward}
  r_{\mathrm{jv}} = \exp \left(
  -5\times10^{-6} \sum_{\mathbf{\dot{q}} > k\cdot \mathbf{\dot{q}_{lim}}}  \| \mathbf{\dot{q}} \| ^2
  \right).
\end{align}

%% full reward function
The full reward function is given by:
\begin{multline}
\label{eq:reward}
  r = w_1 r_{\mathrm{grf}} + w_2 r_{\mathrm{spd}} + w_3 r_{\mathrm{rv}} + \ldots \\
    w_4 r_{\mathrm{av}} + w_5 r_{\mathrm{height}} + w_6 r_{\mathrm{upper}} +\ldots \\
    w_7 r_{\mathrm{posture}} + w_8 r_{\mathrm{jv}},
\end{multline}

\noindent
where, the weights $w_1$, $w_2$, $w_3$, $w_4$, $w_5$, $w_6$, $w_7$, $w_8$ are set to
0.225, 0.225, 0.100, 0.100, 050, 0.100, 0.100, 0.100, respectively.

\subsection{Dynamics Randomization}
Since policies trained in simulation interact with an imperfect representation of the world,
they are prone to overfitting and show brittleness on the hardware. A common approach to
overcome this is to randomize various robot model and environment parameters, such as mass,
intertia, motor strength, latency, ground friction, etc~\cite{2020-rss-RNNCassie, 2021-rss-RMA}.

In our work, we carefully select the variables that are needed to be randomized for a better transfer.
Firstly, we can expect the mass and position of the center of mass (CoM) of each link to be different on
the real system than in the simulation, due to the distribution of electronics and mechanical parts
within a link. We randomize the mass of each link by $5\%$ and randomize the CoM positions by $5cm$ at
the start of each episode during training.

Secondly, prior work shows there exists a significant amount of friction between the motor and the load \cite{cisneros2020reliable}
in geared transimision systems. 
This frictional torque is difficult to identify or even model in simulation. MuJoCo allows the simulation
of static friction and viscous friction. Hence, we randomize the static friction magnitude in the uniform range
of $(2, 8) \SI{}{\N.\m}$ and coefficient for viscous friction in the uniform range of $(0.5, 5) \SI{}{\N\m\per rad.s^{-1}}$,
based on coarse identification performed in \cite{cisneros2020reliable}.

Besides the mass, CoM positions, joint friction, we do not randomize any other robot dynamics parameters
during training.

\subsection{Terrain Randomization}
In order to enable the real robot to walk robustly over uneven terrain, we expose the policy to
randomized terrains during training. In MuJoCo, the terrains are represented using a height fields -
a $2D$ matrix comprising of elevation data. As generating new height fields online during the training
may cause slowness during training, instead, we generate one random height field at the start of the
training and then randomize its relative position to the flat floor in each episode. The flat ground
plane and the height map are simulated simulatenously, so that the floor resembles a terrain with
obstacles of varying heights scattered randomly. In this way the robot is explosed to unevennes of
maximum height of $\SI{3.5}{\cm}$. The introduction of terrain randomization in this work is done
uniformly through the training; but it could also be done gradually according to a curriculum to promote
smoother learning.

\subsection{Simulating Back-EMF}
In order to simulate the phenomenon of poor torque-tracking (observed on the real system),
during the training phase, we introduce a modification to the applied torque for each joint
at each simulation timestep. The modification is implemented by injecting a counter-torque that
scales with the joint velocity, specifically, by using the following equation:

\begin{align}
\label{eq:tautrack}
  \text{tau}_{applied} = \text{tau}_{pd} - \text{k}_{bemf} \cdot \dot{q},
\end{align}

\noindent
where, $\text{tau}_{pd}$ denotes the torque at the output of the PD controller, and $\dot{q}$
represents the joint velocity. The damping coefficient, $\text{k}_{bemf}$, is unknown
for the real system. During training, we randomize
$\text{k}_{bemf}$ to simulate different tracking behavior. The coefficient
for each joint is sampled uniformly within $[5, 40]$ at every $\SI{100}{\ms}$. This
range was determined empirically by comparing the simulation environment and
logs from a real robot experiment.

During training, $\text{tau}_{applied}$ is observed directly by the policy, i.e.,
$\text{tau}_{obs} = \text{tau}_{applied}$.

%% [TODO: add details about training for omnidirectional]

%%%%%%%%%%%%%%%%%%%%%%%%%%%%%%%%%%%%%%%%%%%%%%%%%%%%%%%%%%%%%%%%%%%%%%%%%%%%%%%%

%% file: sections/experiments.tex
%%%%%%%%%%%%%%%%%%%%%%%%%%%%%%%%%%%%%%%%%%%%%%%%%%%%%%%%%%%%%%%%%%%%%%%%%%%%%%%%
\section{Experiments}

\subsection{RL Policy}
\textbf{Training Details.} As in \cite{singh2022learning}, both the actor and critic policies are
represented by MLP architectures to parameterize the policy and the value functions and use
Proximal Policy Optimization (PPO) \cite{schulman2017proximal} for training. Both
MLP networks have 2 hidden layers of size 256 each and use \textit{ReLU} activations. Each episode rollout
spans a maximum of 400 control timesteps (equivalent to $\SI{10}{\second}$ of simulated time), and may
reset if a terminal condition is met. Each training batch holds 64 of such rollouts. The learning rate was
set to 0.0001. We use the \textit{LOSS} method \cite{siekmann2021sim} , which adds an auxiliary
loss term (in addition to the original PPO loss term) to enforce symmetry. Training the FF policy takes around
12 hours to collect a total of 120 million samples for learning all modes, on a AMD Ryzen Threadripper PRO 5975WX
CPU with 32 cores.

\subsection{Implementation on Real Robot}
\label{subsec:real-robot-fb}
We propose to include the actual applied torque in the observation
vector to our RL policy. As mentioned previously, the applied torque
on the real robot is extracted from the measurements of the current
sensors in the motor drivers. The measured current $I_{meas}$ is multiplied by
the torque constant $k_T$ and the gear ratio $g$ corresponding to the joint
and fed to the policy, as follows:

\begin{align}
\label{eq:current_to_torque}
\text{tau}_{mot} = I_{meas} \cdot k_T \\
\text{tau}_{obs} = g \cdot \text{tau}_{mot}
\end{align}

It is important to note that the applied torque here refers to the
torque applied at the level of the actuator scaled to the joint space. The torque applied to
the load (i.e. the robot link) cannot be measured on the HRP-5P
robot due to absence of joint torque sensor. The difference between
the \textit{scaled} actuator torque and the joint torque can in fact be
quite significant due to the presence of static friction and viscous
friction.

The policy is executed on the control PC of the robot (specifications:
Intel NUC5i7RYH i7-5557U CPU with 2 cores, Ubuntu 18.04 LTS PREEMPT-RT kernel),
and is implemented as an \textit{mc-rtc}\footnote{\url{https://jrl-umi3218.github.io/mc_rtc/index.html}},
controller in C++. The inference
is done at $\SI{40}{\hertz}$ with the PD controller running at $\SI{1000}{\hertz}$.
Policy inference is quite fast, taking only around $\SI{0.2}{\ms}$. The
global run of the controller is around $\SI{1}{\ms}$.

%%%%%%%%%%%%%
%%%%%%%%%%%%%
\begin{figure}
  \includegraphics[width=\linewidth]{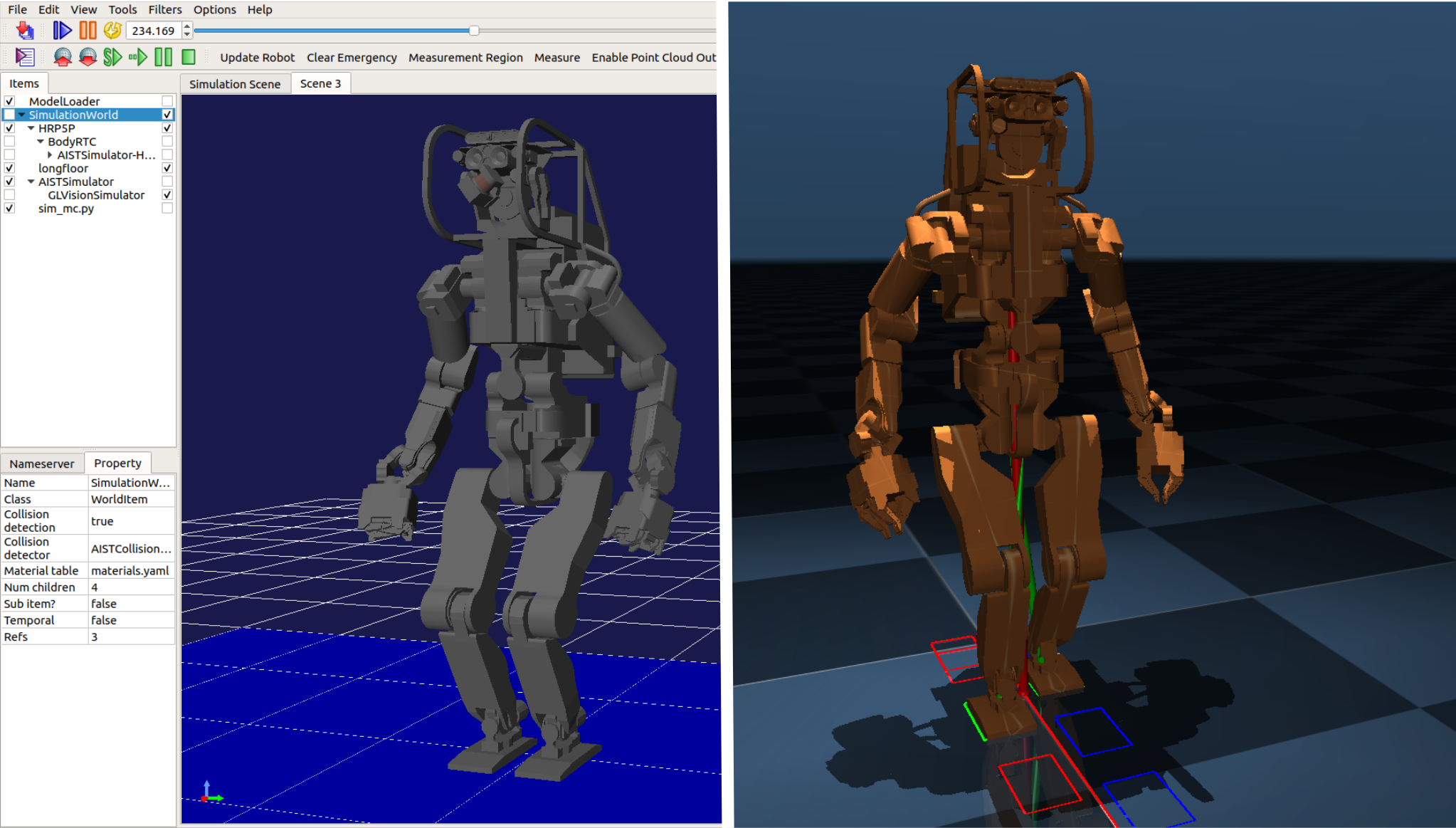}
  \caption{\textbf{Sim-to-sim validation. }Simulating HRP-5P in Choreonoid (left) and MuJoCo (right)
    using the mc-openrtm and mc-mujoco interfaces respectively.}
  \label{figure:sim-to-sim}
\end{figure}
%%%%%%%%%%%%%
%%%%%%%%%%%%%

%%%%%%%%%%%%%
%%%%%%%%%%%%%
\begin{figure}[t]
  \centering
  \includegraphics[width=\linewidth]{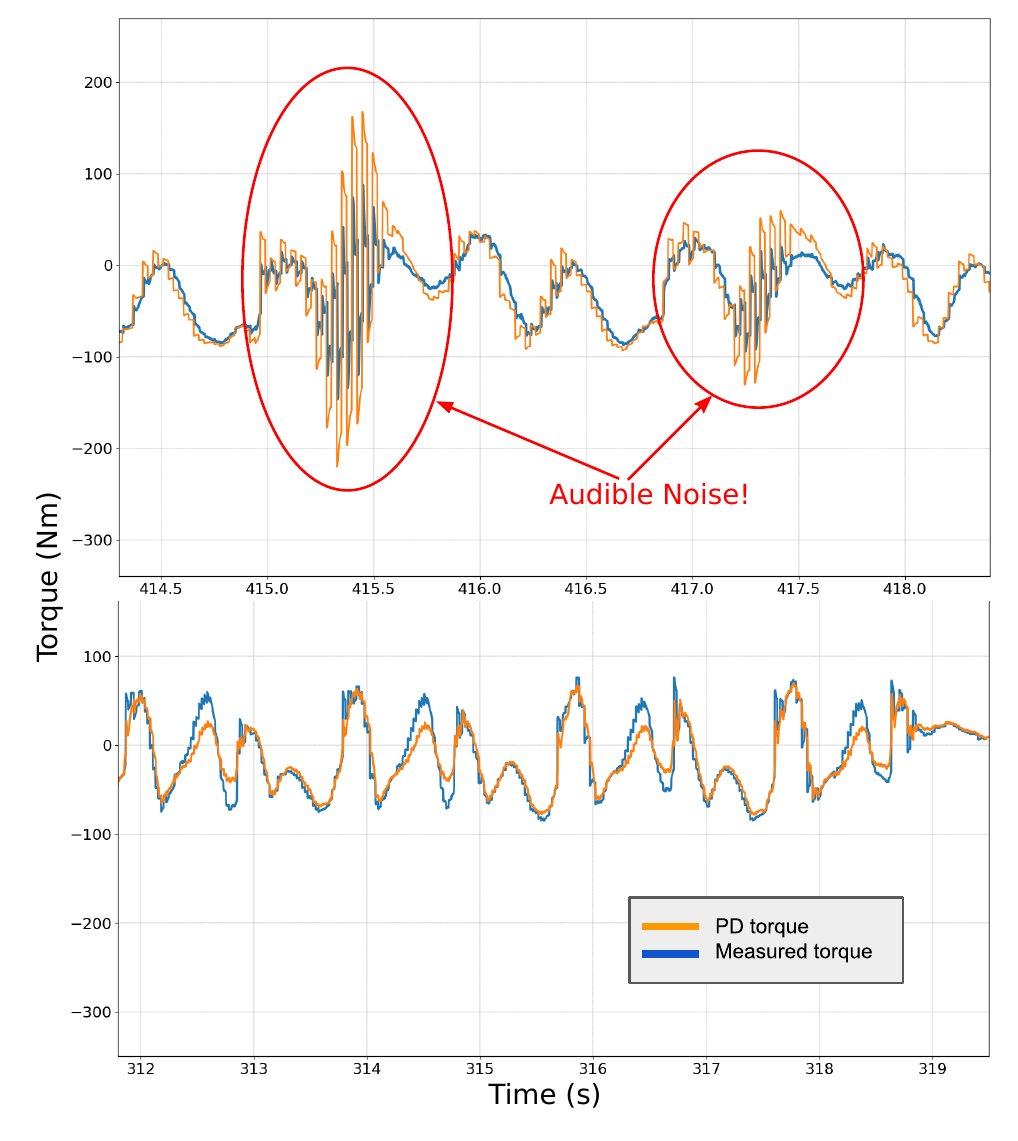}
  \caption{
    \footnotesize \textbf{Baseline (top) v. Proposed policy (bottom): Real experiment logs for torque-tracking
      on the right hip roll joint}. We observe that the torque-tracking becomes
    unstable for the baseline policy (circled) when the robot is turning and an
    audible noise can be heard from the joint. Such effects are not observed with
    the proposed approach, and the robot's swing leg behaviour visually appears to be more stable.
  }
  \label{figure:combined-tracking}
\end{figure}
%%%%%%%%%%%%%
%%%%%%%%%%%%%

\subsection{Sim-to-sim Validation}
While we use MuJoCo as the training environment, we perform thorough evaluations
also in the Choreonoid simulator before real robot deployment.
Choreonoid is traditionally a more popular choice for simulating
humanoid robot controllers \cite{nakaoka2012choreonoid}. We use the \textit{mc-rtc} control framework for executing
the policy onboard the control PC of the robot.

This allows us to evaluate the same controller code transparently in 3 different environment:
(1) in MuJoCo using the \textit{mc-mujoco} interface \cite{singh2023mc},
(2) in Choreonoid and
(3) on the real robot,
both using the \textit{mc-openrtm} interface for communicating with OpenRTM middleware \cite{ando2005rt}
used in the HRP robots and Choreonoid simulator.

We found the contact modelling in Choreonoid to be more stable than the contact
modelling for height fields in MuJoCo. Hence, Choreonoid forms an important part of
our pipeline for evaluating policies on uneven terrain before real robot deployment.
Nevertheless, besides some small differences, we did not observe any major discrepancies
in the policies behavior between MuJoCo and Choreonoid during evaluation.

\subsection{Ablation study}
We perform ablation on the two main ingredients proposed in this work
for \textit{sim2real} success - $(1)$ training with simulated poor
torque-tracking and $(2)$ providing torque feedback to the policy.

Although real robot experiments are expensive and testing of policies
that are prone to failure can be dangerous, developing a test
environment in simulation is not a credible alternative. We found
that policies that succeed in simulation even in very challenging
circumstances (like degraded torque-tracking, uneven terrain, external
perturbations), can still behave undesirably when deployed on the real robot;
indicative of a large and critical reality gap. Hence, it is
important to study the behaviour of the policies on the real system.
We train 3 different policies to analyze the impact of the proposed approach:

\begin{enumerate}
\item \textbf{Policy A} forms the baseline policy. It is trained without simulated poor
  torque-tracking and without observing the current feedback from the
  actuators. When deployed on the real robot, this policy gives the
  worst peformance. The robot is prone to self-collision between the
  feet when the swing leg lands on the ground. This points to the difficulty
  faced by the policy in controlling the real robot's leg swing motion. This
  is because the policy is trained with perfect tracking in the simulation
  environment but is exposed to degraded tracking on the real robot. Further,
  we attempt to train another policy with an additional termination condition
  on the feet distance ($d < \SI{0.2}{\m}$) to promote a wider stance. In this
  case, self-collision is prevented on the real robot but we can observe that
  the torque-tracking becomes unstable in some regions of the motion with an
  audible noise heard from the joints (see \autoref{figure:combined-tracking}).

%%%%%%%%%%%%%
%%%%%%%%%%%%%
\begin{figure}[t]
 \includegraphics[width=\linewidth]{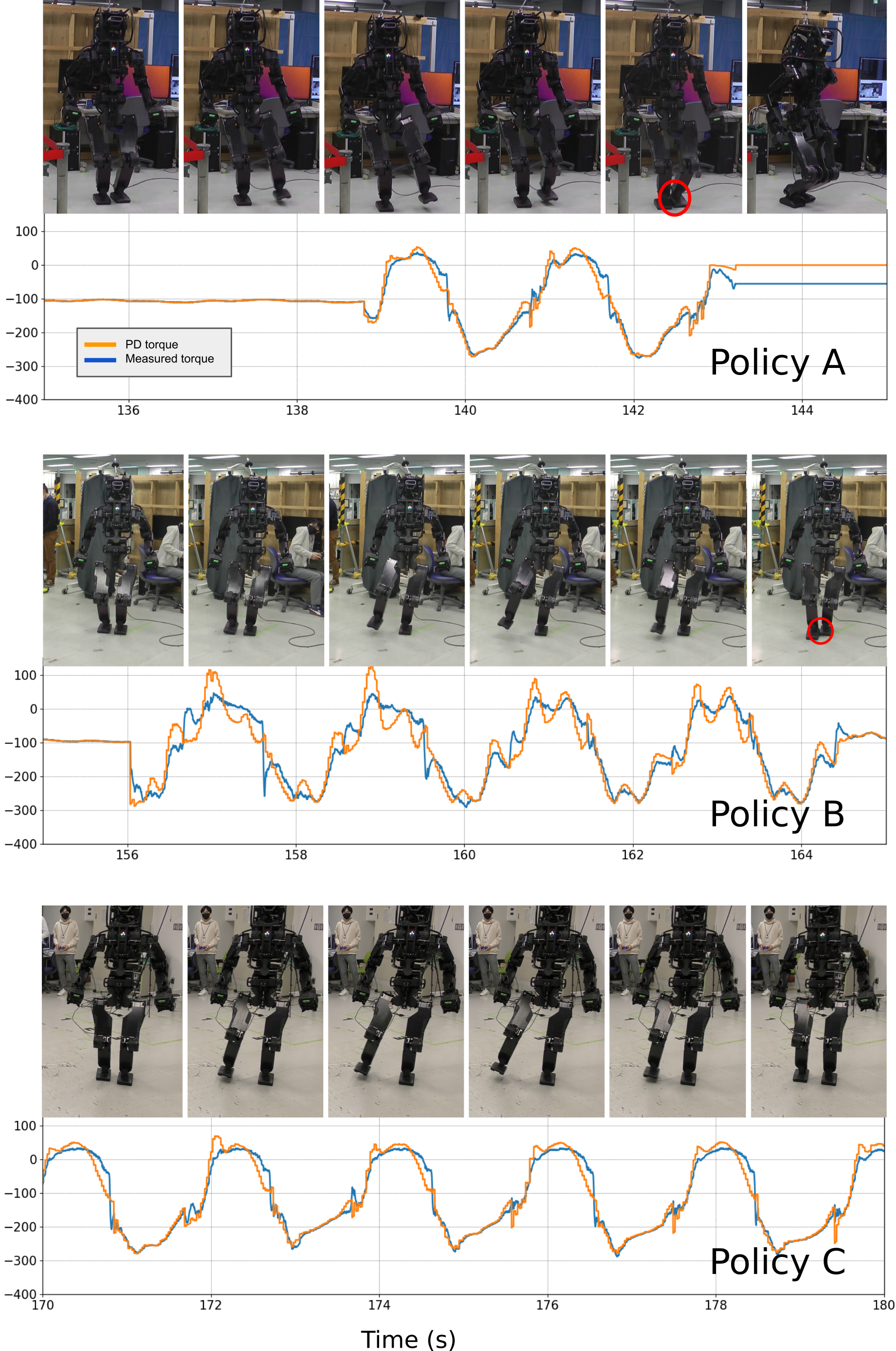}
 \caption{\textbf{Torque-tracking performance on the RKP joint} for 3 policies.
   Policy A is trained without back-emf effect in sim and without feedback. Policy B is trained
   with back-emf but without feedback. Policy C is trained with back-emf and with the torque (current)
   feedback. Policy A and B both lead to self-collision between the feet (circled in red) while
   C is able to perform stable walking.}
  \label{figure:experiment_all}
\end{figure}
%%%%%%%%%%%%%
%%%%%%%%%%%%%

\item \textbf{Policy B} is trained with poor torque-tracking but without torque feedback.
  We replace the inputs corresponding to the torque-feedback with the $\mathbf{0}$
  vector during training and evaluation, while keeping all other parameters the same.
  This policy appears robust in the simulation environment. However, when deployed
  on the real robot, we again observe the self-collision between the feet. The
  speed of the swing leg is also considerably higher, meaning that the policy
  finds it harder to compensate for the changing discrepancy between command
  and applied torque. From this observation, we conclude that providing the torque
  feedback (from the measured current) is vital for the policy to adapt to the
  degraded torque-tracking environment.

\item \textbf{Policy C} is trained using the proposed approach of simulating
  poor torque-tracking plus providing feedback from current measurements to the policy.
  This policy appears significantly more stable on the real robot. Self-collision
  is not observed and there is no audible noise during any part of the motion. The
  robot could successfully walk upto several meters, including turning, stepping
  in-place and standing. The robot could also walk over uneven terrain consisting
  of rigid and soft obstacles upto $\SI{2}{\cm}$ high.
\end{enumerate}

We further analyze the real robot experiment logs corresponding to the 3 policies
in \autoref{figure:experiment_all} for torque-tracking on the ``RKP'' (right knee
pitch) joint. The ``RKP'' joint is chosen because the tracking errors are more
noticeable for this joint. And also because the knee joint is expected to have
a more consequential impact on the walking behavior (than compared to the hip yaw joint, say).
We observed self-collision between the feet in case of Policy A and Policy B, while
Policy C can perform stable walking and handle uneven terrain well. The tracking for
the ``RKP'' joint for A and C looks somewhat similar, however, in case of Policy C,
the feedback allows the policy to react to the tracking error in the previous timestep
and achieve better control on the swing motion. For B, the tracking is observed to be
much worse. Tracking for ``RCR'' (right hip roll) joint for Policy A (retrained for wider
feet distance) and C are also showng in \autoref{figure:combined-tracking}.

Notably, providing the torque feedback will not eliminate tracking errors because
it is difficult for the policy to anticipate the errors in the future. We
believe that incorporating history data in the observation space will also be
beneficial.

\subsection{Comparison to model-based controller}
\label{subsec:robustness-tests}
We compare the robustness of our policy for locomotion on uneven
terrain to an existing model-based controller for humanoid locomotion.
For the test, we use the open-source \textit{BaselineWalkingController}
\footnote{\url{https://github.com/isri-aist/BaselineWalkingController}},
which provides an implementation of walking control based on linear inverted
pendulum mode (LIPM). This method combines online CoM trajectory generation
based on preview control, ZMP feedback based on the divergent component of
motion (DCM) of LIPM, and foot contact force control based on the damping control
law \cite{caron2019stair, kajita2003biped, kajita2010biped}.

Our test environment consists of a stack of padded carpet tiles, each of
thickness $\SI{0.6}{\cm}$, placed on
a flat floor. The robot starts from some distance ahead of the stack
and needs to go across while stepping on top of the stack. Since the
tiles are made from a soft material, the obstacle forms a compliant
support surface - which is more challenging from a balance perspective.
The \textit{BaselineWalkingController} could succeed on a stack of 3 tiles but
failed on a stack of 4 tiles (nearly $\SI{2.4}{\cm}$ in height) --- the robot loses balance and falls when the
support leg is on the stacked carpets. On the other hand, the RL policy trained
with our approach could succeed on a stack of 5 carpets ($= \SI{3}{\cm}$ high)
on 2 of 2 trials. The policy also succeeds in making several partial contacts on the
obstacle, where the foot is placed partially on the stack.
(The tests are shown in the supplementary video.)

While there exist newer model-based approaches for humanoid locomotion that
may provide greater robustness \cite{murooka2022centroidal,romualdi2022online},
our test still provides valuable insights into the robustness of model-free RL
policies against LIPM-based bipedal locomotion controllers. The critical factors
responsible for the higher robustness in the case of controllers based on deep RL
is subject to debate, but we believe the low PD gains, feedback nature of the policy,
and the absence of strict constraints on feet trajectories, may play an important role.

\subsection{Feedforward vs. Memory-based Policies}
\label{subsec:policy-arch}
In this subsection, we study the behavior of feedforward (FF) MLP policies, without any information
about historical states, to that of memory-based policy architectures like Long Short-Term Memory (LSTM)
and FF policies with observation history using simulation and real robot experiments.
Specifically, we analyzed the following 3 types of network architectures:
\begin{enumerate}
  \item[(a)] FF policy (vanilla), consisting of a MLP observing only the current observed state $o_{t}$
  \item[(b)] FF policy with history, which observes the current state along the previous 3 states $\{o_{t}, o_{t-1}, o_{t-2}, o_{t-3}\}$
  \item[(c)] LSTM policy with two hidden layers of 128 units each.
\end{enumerate}

\begin{figure}[t]
  \begin{subfigure}{\linewidth}
    \includegraphics[width=\linewidth]{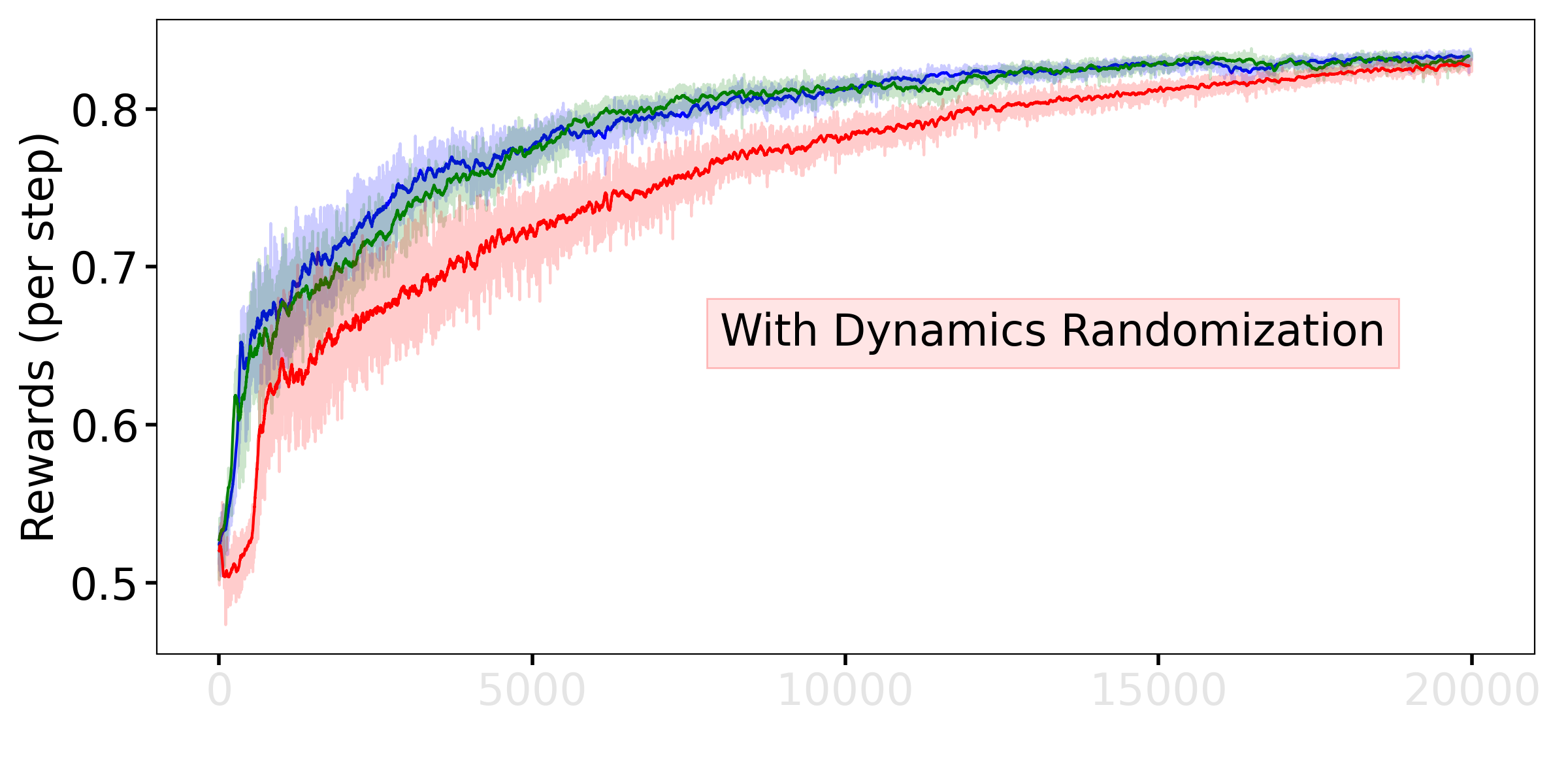}
  \end{subfigure}
  \begin{subfigure}{\linewidth}
    \includegraphics[width=\linewidth]{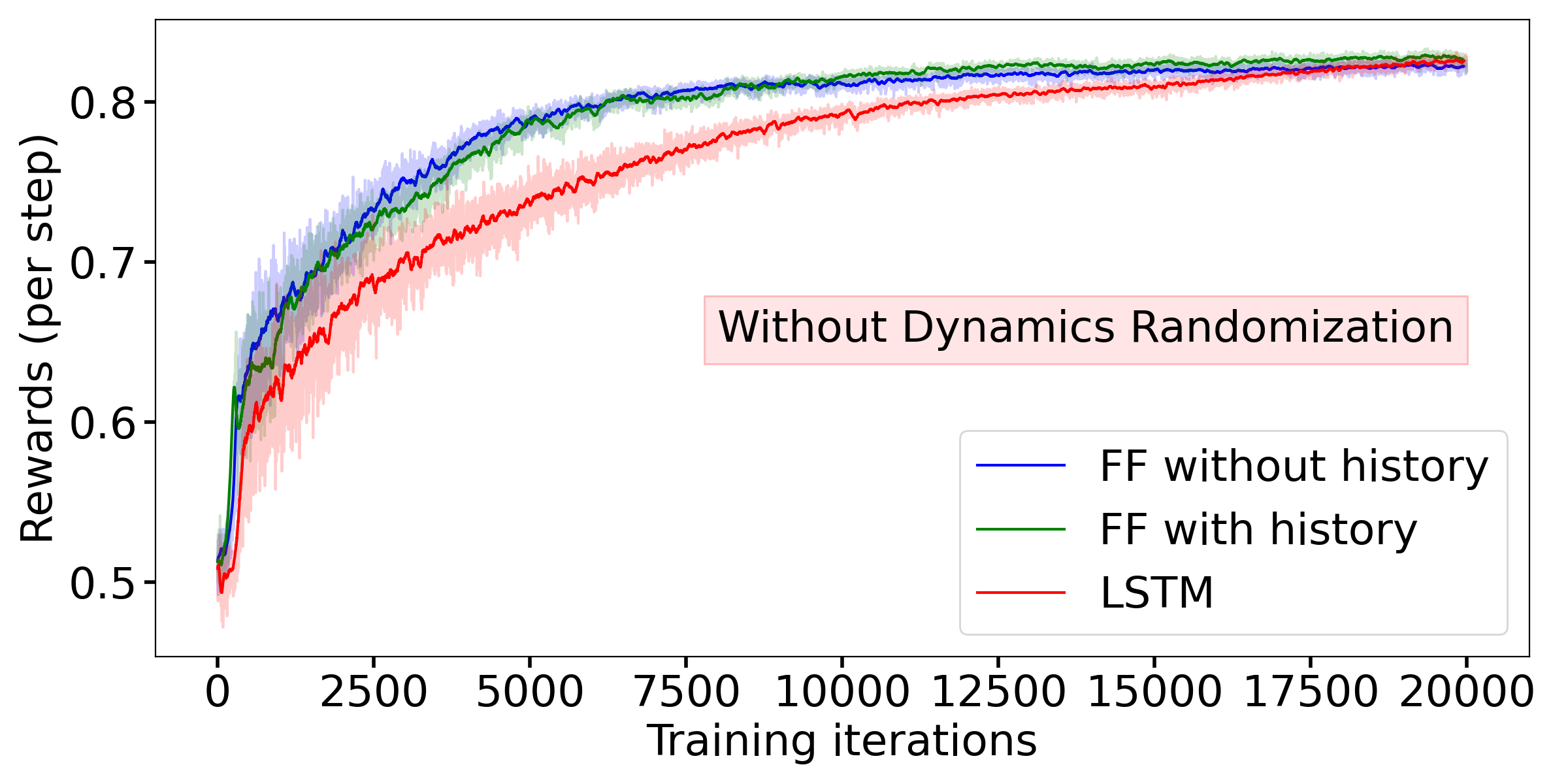}
  \end{subfigure}
    \caption{\textbf{Comparison of Training Rewards} for different policy architectures,
      \textbf{Top:} All policies trained with dynamics randomization. FF policy without history
      (labeled FF (vanilla)) and FF policy with observation history were found to have
      similar reward curves. LSTM policy achieved the lowest rewards in the beginning,
      but appeared to converge roughly to the same level. Both FF policies were tested
      on the real robot, and FF (vanilla) was found to behave most favourably.
      \textbf{Bottom:} Training without dynamics randomization leads to faster convergence and,
      again, LSTM achieves the lowest reward. Policies without dynamics randomization
      were not deployed on the real robot.
    }
  \label{figure:policy_memory}
\end{figure}

Both FF policies have 2 hidden layers of size 256 nodes.
The critic network for each policy had the same architecture as the agent, except for the output layer.
We trained each architecture, with and without dynamics randomization (DR).
None of the policies were provided any information about the dynamics
disturbances. All experiments were performed on a flat and rigid floor.
The reward curves are plotted in \autoref{figure:policy_memory}.

During training, we found the FF policies to slightly outperform the LSTM policies in terms of
reward collection. This is true even when policies were trained without dynamics randomization,
but unsurprisingly, all policies appeared to converge faster with no DR. Vanilla FF policy and FF policy
with history had nearly overlapping rewards curves. We stopped the training after 20000 iterations,
but it is possible that the LSTM reward may eventually surpass FF policies due to overfitting to the
simulation dynamics.

%%%%%%%%%%%%%
%%%%%%%%%%%%%
%%%%%%%%%%%%%
%%%%%%%%%%%%%

FF policies trained without DR were not deployed on the real robot after initial unfavorable results with
prototype policies. The swing leg motion visually appeared to be different from the simulated robot and
the robot was prone to slight instability. However, we did not observe shakiness or extreme instability
even with no DR.
With DR, the FF policy with observation history seemed to perform similar
to the vanilla FF policy for stepping in-place on the real robot.
However, while walking forward and turning we noticed the former to suffer
from gradual degradation --- gait symmetricity appears to break and body oscillations appear to grow.
We hypothesize that this is due to the accumulation of \textit{sim2real}
errors with time, that would otherwise not occur if the network only observed the current state $o_{t}$.
We did not deploy the LSTM policy on the hardware as its slower inference speeds may prohibit
real time execution. We expect it behave similar to FF policy with history.
The vanilla FF policy trained with DR performed the best in all real robot experiments, and hence,
emerged as our preferred choice for policy architecture.

Our result appear to be somewhat in contrast to the prior findings for Cassie \cite{2020-rss-RNNCassie}
that report effectiveness of memory-based networks for achieving \textit{sim2real} success. We believe
this is due to the narrow range of randomization performed during training for our robot, meaning
that the state-history compression ability of memory-based agents provided no added benefit.
Since training the LSTMs takes nearly 3 times as long as the FF policies (36 hours compared to 12
hours for FF), our results show the potential advantage of using FF policies with targetted
dynamics randomization for \textit{sim2real} methods. We also note the importance of an accurate
initial model of our robot for the same.
%%%%%%%%%%%%%%%%%%%%%%%%%%%%%%%%%%%%%%%%%%%%%%%%%%%%%%%%%%%%%%%%%%%%%%%%%%%%%%%%

%% file: sections/conclusion.tex
%%%%%%%%%%%%%%%%%%%%%%%%%%%%%%%%%%%%%%%%%%%%%%%%%%%%%%%%%%%%%%%%%%%%%%%%%%%%%%%%
\section{Discussion \& Conclusion}

In this work, we developed a system to train control policies for a life-sized humanoid robot HRP-5P.
We identified that the main \textit{sim2real} gap for these types of large robots arises from poor torque-tracking
of the motor control systems due to high gear-ratio. We simulated back-EMF and applied torque feedback to the policy
to combat the \textit{sim2real} gap. Policies were trained in simulation and directly transferred to the hardware.

Our experiments show that providing the current feedback is a key ingredient for reliable
\textit{sim2real} transfer. Without the proposed feedback signal, the policy is prone to
failure in controlling the leg swing motion, often causing self-collision between the legs.
We could not achieve \textit{sim2real} success without simulating poor torque-tracking during
training. For robots with joint-level torque sensors, we believe our proposed approach can
yield better performance by accounting for the frictional torque in the joints.

We also provide ablation analysis on the need for memory-based policy architectures. Our
results show that a feedforward MLP could be sufficient for successful transfer of policies
learned in simulation, in cases where dynamics randomization is performed in a narrow
range. Since memory-based networks like LSTMs can be more compuationally expensive to train
and are prone to overfitting, for this work, we chose MLP policies for real robot deployment.

We could achieve zero-shot transfer without performing aggressive manual tuning
of the reward function or randomizing dynamics variables to wide, unreasonable ranges (an often
omitted part from the literature). It points to the potential effectiveness of an accurate robot
model for training as well as careful identification of key factors responsible in overcoming
the \textit{reality gap}. However, the policy appears to make large swaying motion of the swing
leg, which could be eliminated by reducing the PD gains of the hip
roll joints and relaxing the coefficient of the penalty on joint velocity reward term.

We compared the RL policy to a conventional model-based approach for bipedal locomotion on
the real humanoid platform and obtained encouraging results. The RL policy could handle obstacles
over $\SI{3}{\cm}$ while the robot lost balance and falls with the model-based controller for
obstacles over $\SI{2}{\cm}$. Although there are newer model-based methods that can tackle
larger obstacles, our tests provide promising evidence in the favor of compliant joint tracking,
closed-loop control and relaxed trajectory constraints --- offered by the RL approach.
Visually, the robot exhibits a cleaner and efficient gait under the model-based controller as
opposed to a excessive sway under the RL policy. In the future, this could be tackled through
the use of reference motions and fine-tuning of learning parameters.

We release the source code for RL training and evaluation in MuJoCo for reproducibility
\footnote{\url{https://github.com/rohanpsingh/LearningHumanoidWalking}}.

\textbf{Future work.}
In the future, we plan to expand the framework for developing a policy for \textit{backwards}
locomotion and tackle even more challenging terrain. We also hope to identify and overcome
other factors inhibiting better \textit{sim2real} transfer.
%%%%%%%%%%%%%%%%%%%%%%%%%%%%%%%%%%%%%%%%%%%%%%%%%%%%%%%%%%%%%%%%%%%%%%%%%%%%%%%%

%% file: sections/acknowledgements.tex
%%%%%%%%%%%%%%%%%%%%%%%%%%%%%%%%%%%%%%%%%%%%%%%%%%%%%%%%%%%%%%%%%%%%%%%%%%%%%%%%
\section*{Acknowledgements}
The authors thank all members of JRL for providing their support in conducting robot
experiments that were done during the production of this work. We are especially grateful
to Hiroshi Kaminaga, Mitsuharu Morisawa, and Mehdi Benallegue for the many insightful
discussions. This work was partially supported by JST SPRING Fellowship Program,
Grant Number JPMJSP2124.
%%%%%%%%%%%%%%%%%%%%%%%%%%%%%%%%%%%%%%%%%%%%%%%%%%%%%%%%%%%%%%%%%%%%%%%%%%%%%%%%